# Reconfigurable drone system for transportation of parcels with variable mass and size

Fabrizio Schiano[1,2], Przemyslaw Mariusz Kornatowski[1,2], Leonardo Cencetti[1], Dario Floreano[1], *Senior Member, IEEE*[1]

*Abstract*—Cargo drones are designed to carry payloads with predefined shape, size, and/or mass. This lack of flexibility requires a fleet of diverse drones tailored to specific cargo dimensions. Here we propose a new reconfigurable drone based on a modular design that adapts to different cargo shapes, sizes, and mass. We also propose a method for the automatic generation of drone configurations and suitable parameters for the flight controller. The parcel becomes the drone's body to which several individual propulsion modules are attached. We demonstrate the use of the reconfigurable hardware and the accompanying software by transporting parcels of different mass and sizes requiring various numbers and propulsion modules' positioning. The experiments are conducted indoors (with a motion capture system) and outdoors (with an RTK-GNSS sensor). The proposed design represents a cheaper and more versatile alternative to the solutions involving several drones for parcel transportation.

## I. INTRODUCTION

CURRENTLY, aerial transportation of parcels of different mass, size, and shape requires drones with different sizes, structures, and payload bay designs [1]–[3]. Unmanned Aerial Vehicles (UAVs) can come in various shapes and configurations, broadly categorized into two main groups: fixed- and rotary-wing vehicles. Each group has its strengths and weaknesses that lead them to be used in different applications.

In this work, we are interested in solutions involving multicopter vehicles because of their lower complexity, higher maneuverability, and lower manufacturing cost. The range of applications of a single drone is limited by the maximum lifting capabilities of its propulsion system.

A possible solution consists in using multiple drones to lift and carry heavy cargo [4]–[8]. However, this approach requires complex control algorithms for robot coordination and communication. Secondly, combining many drones to deliver heavy parcels increases the use of similar/redundant components (e.g., autopilot, external computer, sensors, airframe), which unnecessarily increases the system's mass, reducing its payload, and range capabilities.

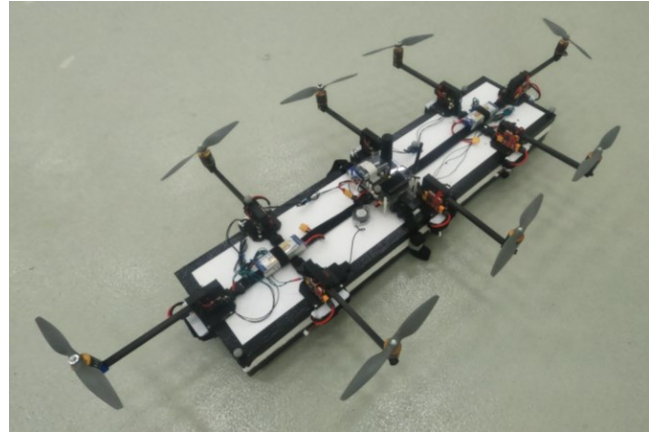

Fig. 1. A reconfigurable drone based on a modular design. The drone is composed of eight propulsion modules and one computation module attached to a parcel. The propulsion modules are attached in pairs (one in front of the other) to create a symmetrical octocopter airframe morphology. The parcel is simulated by a Styrofoam block (100 x 25 x 8 cm) that weighs 420 grams.

State of the art presents several articles about parcel transportation using multicopter UAVs. A complete survey is out of the scope of this work (the interested reader is referred to Villa et al. [9] for further details). We classified the literature in three categories (Fig. 2): (A) single multicopters [1]–[3], (B) multiple multicopters [4]–[8], [10] (C) modular multicopters [11] [12]. The first two categories (A and B) can be split into two subcategories: rigidly attached, where the interface between the drone frame and the parcel is rigid (A1, B1) [4], [13]–[17], and cable suspended, where the interface is represented by a cable (A2, B2) [6], [8], [18]–[25]. This subcategorization does not apply to the last category (C) since a propulsion module with a single motor and propeller cannot be connected through a cable.

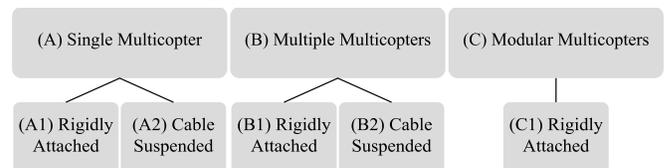

Fig. 2. Classification of the literature on multirotor UAVs used for parcel transportation.

Manuscript received: April 3, 2022; Revised: June 30, 2022; Accepted: August 23, 2022.

This letter was recommended for publication by Pauline Pounds upon evaluation of the reviewers' comments.

[1]This work was supported by the Swiss National Science Foundation through the National Centre of Competence in Research (NCCR) Robotics and Armasuisse. (Corresponding author: Fabrizio Schiano) The authors are with the Laboratory of Intelligent Systems, School of Engineering, Ecole Polytechnique Federale de Lausanne, Lausanne CH1015, Switzerland (e-mail: fabrizio.schiano, przemek.kornatowski, cencetti.leonardo@gmail.com; dario.floreano@epfl.ch). Fabrizio Schiano is also with Leonardo S.p.A., Leonardo Labs, 00195 Roma.

[2] Fabrizio Schiano and Przemyslaw Mariusz Kornatowski contributed equally to this work.

Digital Object Identifier (DOI): see top of this page.

This system paper proposes a reconfigurable drone based on a modular design that adapts to different cargo shapes, sizes, and masses (Fig. 1). Moreover, we introduce a method for the automatic generation of drone configurations and suitable autopilots. In this work, with "reconfigurable drone," we mean a drone comprised of a single computation module and several propulsion modules which can be easily attached to the edge of a parcel. The propulsion module includes single motor, propeller, battery, and electronics to control the motor. The module can be multiplied to reach the lifting requirements of the transported parcel. The computation module is the system's brain, which controls all the connected modules via cables. In this solution, the parcel becomes the body of the multicopter drone. This approach also requires simpler control to fly compared to an approach with multiple drones. Finally, a reconfigurable modular drone would have lower scalability and repair costs [26] since modules are simpler and cheaper than drones with comparable lifting performance. Additionally, the modular approach allows users to replace a broken module and quickly launch the system.

There has been recent interest in modular drones. Indeed, both a seminal work by Oung et al. [11] and a recent pioneering work by Mu et al. [12] already presented this idea. However, in previous work, the user had to decide the number of modules and their arrangement. Our method includes an algorithm for defining the appropriate configuration based on the mass and dimension of the cargo.

Compared to [11], [12], which showed only hovering experiments, we successfully tested our drone in indoor and outdoor environments. The experiments involved multiple flights, transporting parcels of different masses and sizes. We chose a Lissajous flight trajectory to excite all the degrees of freedom of the drone (indoor tests) and point-to-point GNSS-based navigation (outdoor tests).

## II. Hardware

The proposed drone system uses a centralized control architecture to avoid redundant components and unnecessary mass. The system comprises two types of modules: a central computation module (Fig. 3a) and a propulsion module (Fig. 3c).

The central module is a computation and communication unit composed of: a Pixhawk 4 autopilot, an NVIDIA Jetson Nano companion computer, a cable roller, a magnetic connector, Wi-Fi and RF (433 MHz) dongles, an RC receiver, a 4G modem, and an RTK GPS receiver (the last two modules are only used for outdoor flights). Moreover, the computation module has four rigidly attached clips that allow fixing the module to a parcel with two belts (Fig. 4). Each propulsion module includes a motor, a propeller, a UAVCAN Electronic Speed Controller (ESC), a UAVCAN Power Management Board, a cable roller, a battery, a magnetic connector, and a mechanical attaching interface (Fig. 5, Fig. 6). The ESCs and the Power Management Boards use the UAVCAN protocol for signal distribution among all the modules via Daisy-Chain architecture (serial interface), which simplifies cable distribution and prevents a tangle of cables. Additionally, cable rollers were used to adjust the cable length between modules to prevent cable entanglement. The positioning of the battery in each propulsion unit (instead of only in the central module) enables the use of battery types with parameters (e.g., battery capacity and discharge rate) adapted to the specificities of the propulsion module. A UAVCAN Power Management Board in each propulsion module informs the autopilot about each battery's voltage and power consumption. We modified the PX4 firmware to read, log, and report the data from the UAVCAN Power Management Boards to provide this feature. We tested it with 16 boards, but theoretically, the CAN interface allows us to connect 128 modules. Moreover, we implemented an algorithm that triggers safety features according to the number of discharged batteries in the propulsion modules based on the information received from the UAVCAN Power Management Boards.

The propulsion module has a mechanical attaching interface (Fig. 6) to connect the module onto the parcel. The propulsion modules are constrained to be paired with another propulsion module on the opposite side of the parcel to simplify the software algorithm responsible for generating the drone morphology. The modules are fixed to the parcel by a plastic belt attached to the mechanical attaching interface by integrated plastic clips (Fig. 6). In the future, another hardware interface to attach modules could be introduced based on the material of the parcel. For example, in the case of wooden boxes, the propulsion modules could be attached through screws; in the case of plastic boxes, modules could be attached by Gecko-based adhesion tapes, or part of the parcel structure could be used to attach a module, e.g., specifically formed parcels edge.

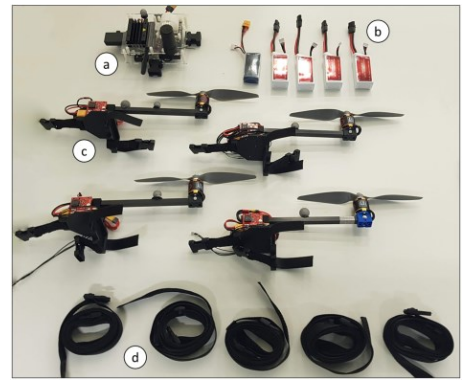

Fig. 3. Reconfigurable drone kit composed of: (a) central module, (b) five batteries, (c) four propulsion modules, and (d) five polyester belts to attach modules to a parcel.

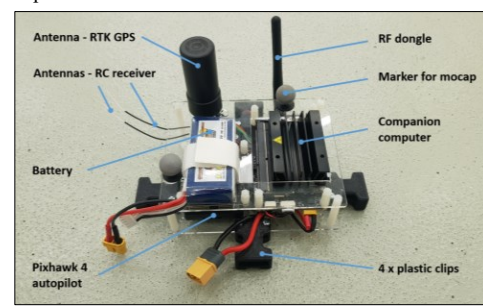

Fig. 4. The central module with described components.

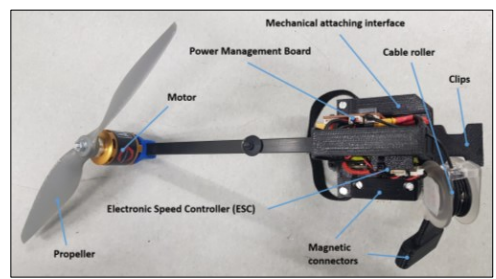

Fig. 5. The propulsion module with described components.

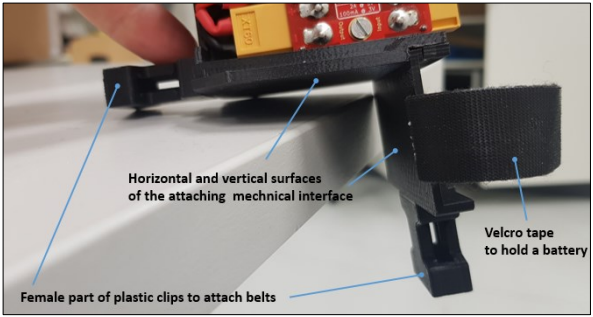

Fig. 6. The mechanical attaching interface of the propulsion module.

## III. Software

We designed a software suite able to assist the user in the process of assembly of the drone. Moreover, the software autonomously chooses the control parameters of the modular drone. The workflow of the modular drone assembly and tuning is shown in Fig. 7. The following subsections will describe its building blocks more in detail.

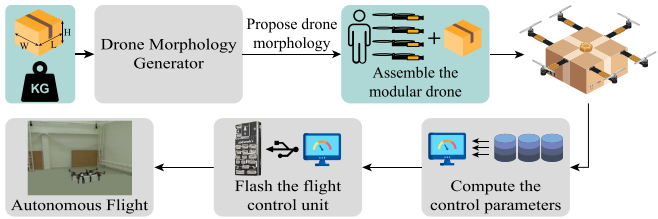

Fig. 7. Workflow of the reconfigurable drone assembly and testing. First, the user measures the physical properties of the parcel to be transported, and the morphology generator uses these parameters to create and propose a suitable configuration. The user is then requested to assemble the drone. Afterward, the control parameters are automatically retrieved from a database built entirely through simulations. The control parameters, the properties of the parcel, and the drone's morphology allow the software to build a custom firmware that is flashed on the central flight control module. The green blocks represent tasks users perform, while gray-colored blocks represent tasks autonomously completed by the software.

### A. Generation of autopilot firmware and parameters tuning for each drone morphology

We developed software to build a simulation environment where different drones can be created programmatically. This environment allows to automatically tune the control parameters of the drones based on their configuration and populate a database with the pairs "drone morphology-controller parameters" Fig. 7.

The software is based on the Gazebo simulator and runs the PX4 Software-In-The-Loop (SITL) stack. This choice allows us an almost seamless transition from simulation to real-world experiments. More specifically, the software takes as inputs:

- a *layout file* (in ".csv" format). This file contains the positions of the motors with respect to the center of the drone (i.e., w.r.t. the IMU position). The generation of this file is described in Section IV.B;
- the mass of the parcel.

Based on these two inputs, it generates:

- a drone *geometry file* (".toml" format). The file describes the propulsion modules' positions and the propellers' spinning direction. PX4 uses the *geometry file* to generate a *mixer file* (".mix" format) to map the outputs of PX4 rate controllers to specific motors.[2] Once the mixer is created, it is added to the PX4 firmware that is then rebuilt into a *custom firmware file* (".px4" format);
- a SDF *model object file* (".sdf" format). The Gazebo simulator interprets this file to create the drone object;
- a ROS *launch file* (".launch" format). This file contains all the information for ROS to launch a Gazebo-PX4 simulation to tune the control parameters of the reconfigurable drone automatically.

Regarding the ROS *launch file*, one of the novelties of our work is the ability of our software to automatically tune the roll and pitch attitude rate PID controllers of the simulated reconfigurable drone. Yaw can notoriously be an issue on modular structures, especially for morphologies that have a significant difference between their two main dimensions (i.e., width and length) [27], [28]. A relatively simple solution to this problem is to tilt the propellers of each module of the vertical axis [28], [29]. A more sophisticated solution - mechanically and mathematically - to mitigate this problem has been used in [30] for a multiple-multicopters setup. However, this work focuses on the whole reconfigurable drone system and not its control. Therefore, we decided to simplify our approach by assuming that we can always use the default PX4 gain for the yaw rate controller in simulation and the real world.

The tuning procedure entails the creation of a simulated test rig in the Gazebo simulator. The software uses the test rig to let the modular drone rotate around one axis (e.g., roll) and block its rotation around another (e.g., pitch). Then, we used the well-known Ziegler-Nichols heuristic algorithm to find some PID gains which allow the drone to have a stable roll/pitch response [31]. This method allows starting from a proportional gain equal to zero. Then, the software slowly increases the proportional gain while analyzing the drone response. The procedure stops when the "Drone Response Analysis" detects an oscillation (Fig. 8). At this point, both the integral and derivative gains are inferred from (1) the current value of the proportional gain (usually called "ultimate gain") and (2) the oscillation period.

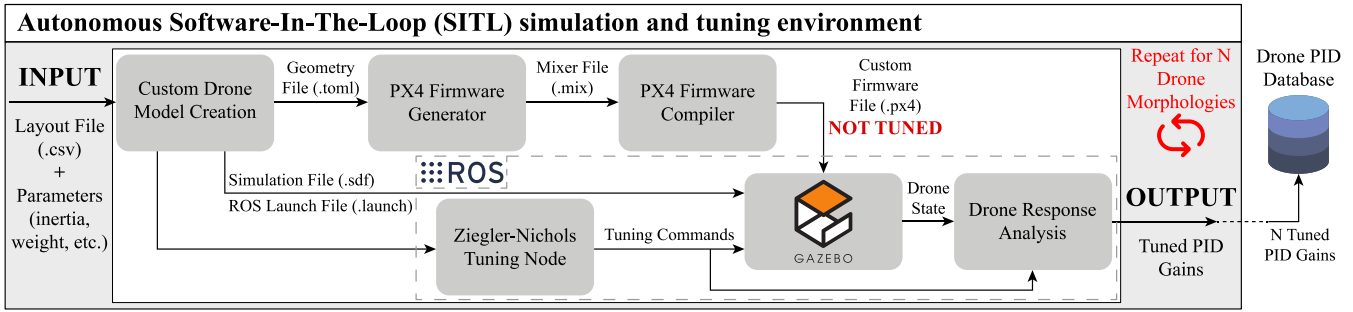

Fig. 8. Workflow of the autonomous Software-In-The-Loop (SITL) simulation and tuning environment for different drone morphologies.

---

[2] https://docs.px4.io/master/en/concept/geometry_files.html

The autotuning procedure described above is used to tune the controllers of several drone morphologies. This step allows building a database in which each drone morphology is associated with its tuned control parameters Fig. 7. The use of this database will be explained at the end of Section IV.B.

### B. Drone morphology generator

Since the parcels to be transported can have different dimensions and masses, the *layout file* of each reconfigurable drone can be different, and it must be generated accordingly. Therefore, we developed a Graphical User Interface in which the user can enter the dimensions and mass of the parcel.

Regarding the inertia, we assume that: (1) the inertia tensor is a diagonal matrix, with the off-diagonal terms equal to zero, and (2) the material of the parcel is known. The current approach is limited to parcels with a diagonal inertia matrix. Lifting this assumption is left as future work. Beforehand, we hardcoded in the software the main parameters of the modules (e.g., mass, thrust, size, and inertia). Based on both the properties of the propulsion modules and the parcel, the software proposes a viable drone morphology (i.e., how many modules are needed and where to place them on the parcel). The morphology is essential to maximize the range and controllability of the modular drone. Indeed, without it, the user would probably place the modules in a suboptimal configuration that could lead to excessive stress and possibly failure of specific motors. Let us denote with:

- $e_H$ the effort at hovering. This parameter is defined as the percentage of the maximum available thrust of all the modules used to keep the drone hovering;
- $s_L$ the space left to mount the modules on the parcel. $s_L$ is computed as the sum of the spaces left between the propellers. For example, in the top-left drone morphology depicted in Fig. : $s_L = \sum_{i=1}^{8} s_i$

Below are the steps to generate a drone morphology. The procedure is an optimization problem with several constraints that will be detailed below:

**Step 1:** The human provides to the software the parcel properties. Then, based on the properties of the propulsion modules available, the software computes all viable configurations in terms of number of modules. Among all the possible configurations, the software chooses only the ones which satisfy the following two criteria: (1) $e_H > e_{H_{th}}$: the software does not consider any configuration that has an effort at hovering greater than $e_{H_{th}}\%$ (in our experiments $e_{H_{th}}$ was set to 60%; this assumption ensures to have at least 40% of the total thrust available for controlling the drone); (2) $s_L > 0$: the software does not consider configurations entailing a number of modules that would not fit on the parcel. The latter criteria is based on the knowledge of the properties of the available modules (i.e., how much space is needed for each module to be mounted on the parcel).

**Step 2:** For all the viable configurations computed in the previous step, the software computes the following score:

$$s = w_e e_H + w_s s_L$$

where $w_e, w_s$ are scalar weights that can be adjusted according to the importance the user wants to give to the effort at hovering $e_H$ and to the space left around the drone $s_L$ respectively. Then, among all the viable configurations computed, the software selects the configuration that achieves the highest value of the score $s$ defined above.

**Step 3:** Once the number of modules is chosen, their distribution around the parcel is calculated by looking at the resulting ratio $r$ between the moments around the $x$ and $y$ axes (when the modules are attached to the parcel). In this step the user has two options. The first is to make sure that the resulting $r$ is optimized to be as close as possible to the following ratio:

$$r_{size} = \frac{p_W}{p_W + p_L}$$

where $p_W$, $p_L$ are respectively the width and length of the parcel. If the user chooses to optimize w.r.t. the ratio $r_{size}$ the distribution of the modules will favor only a homogeneous distribution of the modules around the parcel. The second alternative is to optimize w.r.t. the following ratio (this is the option adopted throughout the experiments presented in Section IV):

$$r_{inertia} = \frac{p_{I_{XX}}}{p_{I_{XX}} + p_{I_{YY}}}$$

where $p_{I_{XX}}$, $p_{I_{YY}}$ are respectively the inertia of the parcel around the $x$ and y axes where x and y are respectively the roll and pitch axes. The choice of the ratio $r_{inertia}$ distributes the modules to have an inertia of the whole drone best distributed around the $x$ and y-axis. If this option is selected, the software will automatically compute the inertia of the whole drone based on its knowledge of the parcel and of the selected modules.

**Step 4:** Based on all the considerations of the previous steps, the software computes the modules' positioning taking into account also the following aspects. First, the modules are positioned by ensuring that the spinning directions are alternating to ensure a zero net torque induced by the propellers (about the vertical axis of the drone). Therefore, the software proposes a modular drone morphology that allows achieving an even load distribution among the motors, minimizing their risk of damage and thus increasing the overall flight time of the drone.

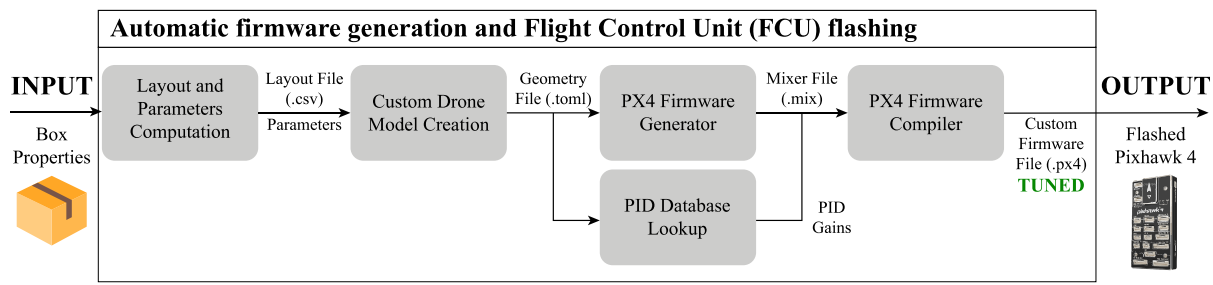

Fig. 9. Automatic Firmware Generator and Flight Control Unit (FCU) Flashing.

**Step 5:** Based on the proposed morphology and on the inertia of the parcel, the software examines the database created as described at the end of Section IV.A and selects the gains of the controller. When the parcel properties are not precisely within the data points present in the database, the software interpolates between existing data points using weighted Gaussians based on the Gaussian Mixture Model (GMM) theory. The direction of flight is also automatically selected in this step. It is chosen by letting the drone have a forward flight direction that exposes the smallest parcel surface (among the two) [32]. Then, thanks to both the mixer files and the PID gains availability, the software builds a custom firmware that already contains the tuned control parameters. Then, the firmware is automatically flashed on the flight control unit (FCU) Fig. 9.

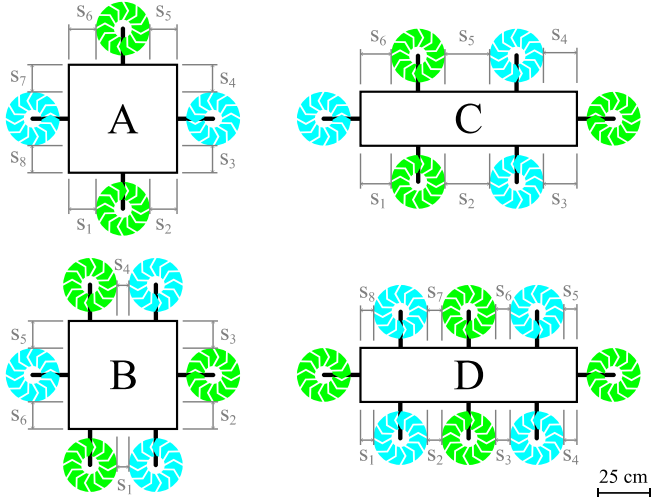

Fig. 10. Schematic representation of the reconfigurable drones with different morphologies used during the experiments. Clockwise and counterclockwise propellers are represented respectively by green and cyan arrows in circles. Dimension lines and arrows show free spaces between the propellers, which are computed during the positioning of modules.

**Step 6:** Finally, the user assembles the drone following the instructions provided by the software written in a text file (i.e., for each module, it is specified its spinning direction and its position w.r.t. the center of the parcel).

After these steps, the reconfigurable drone is ready to fly.

## IV. EXPERIMENTS

We validated the hardware and software described in the previous sections with several flight tests (Fig. 13 – 14, and supplementary video), both indoors and outdoors. Four different morphologies (Fig. 10) were assembled using two rectangular foam shells whose physical properties are listed in Table 1. All platforms were generated, assembled, and automatically programmed with the procedure described above and were able to fly successfully.

Table 1: PHYSICAL PROPERTIES OF THE PARCELS USED FOR THE EXPERIMENTAL PLATFORMS

| Platform | Parcel | Mass | Length | Width | Height |
|---|---|---|---|---|---|
| A | I | 420 g | 0.50 m | 0.50 m | 0.08 m |
| B | II | 760 g | 0.50 m | 0.50 m | 0.08 m |
| C | III | 420 g | 1.00 m | 0.25 m | 0.08 m |
| D | IV | 760 g | 1.00 m | 0.25 m | 0.08 m |

### A. Indoor experiments

The modular drones were tested in a motion capture system arena (10 x 10 x 6 m) equipped with 24 OptiTrack cameras, which provide sub-millimeter accuracy and enable control of the position of the drones in a closed loop. Each drone was equipped with multiple reflective markers (Fig. 4, Fig. 5), allowing its 3D position, and heading determination. These measurements were then transmitted from the ground station to the companion computer (a Jetson Nano running Ubuntu 18.04 and ROS Melodic) on the drone via a Wi-Fi connection (2.4GHz) at a rate of 60 Hz. The transmission and frame transformation introduce a ~50 ms delay in the ground-truth data, which is considered by the PX4 autopilot.

*1) Manual flights*

After assembly, the qualitative performance of each morphology was manually tested (with the Remote Controller (RC)) by taking off and verifying the response to roll, pitch, and yaw commands. This step was necessary to confirm nominal operations of the custom firmware and capture any critical misbehavior before proceeding with further experiments. All platforms were capable of stable take-off, flight, and land, and they did not require any additional intervention.

*2) Hovering flights*

As an intermediate step, each morphology was tested in position control mode, with position and attitude feedback from the OptiTrack system. As for the previous step, the drones were manually controlled during take-off and then transitioned to the autonomous position control mode (PX4 Position Mode). The main goal of this step was to detect oscillatory modes (possibly a symptom of sub-optimal PID tuning). None of the platforms tested suffered from this problem.

*3) Trajectory tracking flights*

Lastly, a pre-programmed flight trajectory was executed by configuring the autopilot in the PX4 offboard mode. We chose a 3D Lissajous trajectory with a single crossing (Fig. 13), enclosed in a 4 x 2 x 2 $m^3$ volume. In contrast with [10], which limited indoor testing to hovering flights or simple A-to-B straight trajectories, our testing strategy provides a complete drone performance evaluation since it excites multiple modes simultaneously. Additionally, complex trajectories can highlight sub-optimal PID tuning. Moreover, these trajectories can provide an effective method to validate our PID tuning and gain-sampling procedure. The testing procedure was fully automated with the implementation of a Python ROS node capable of arming the drone and controlling its position and attitude. Manual switching of the autopilot to offboard mode was done through the Remote Controller (RC) to ensure supervision.

Each flight consisted of five distinct steps, listed in Table 2: starting from the local position [0, 0, 0], (1) the drone was instructed to take-off to an altitude of 2 m; and (2) hold position to ensure consistency in the starting position; then, (3) a single loop of the Lissajous trajectory was performed, controlling both position and heading (i.e., drone always heading tangent to the curve); (4) lastly, the drone was instructed to hover 2 m above the home position; and (5) finally land. After each experiment, the flight log was retrieved and archived for further analysis.

Table 2: INDOOR EXPERIMENT OUTLINE

| Step | Action | Duration |
|---|---|---|
| 1 | Take-off | 5 s |
| 2 | Hold position | 5 s |
| 3 | 3D Lissajous eight (1 loop) | 40 s |
| 4 | Hold position | 5 s |
| 5 | Land | 5 s |

Each experiment lasted 60 seconds from take-off to landing and was repeated nine times for each of the four platforms. The batteries of all modules were replaced every three flights, well before complete depletion, to maximize correlation across experimental results. The quantitative performance of each platform was evaluated by computing the average Root Mean Square Error (RMSE) of the position and yaw with respect to the setpoint. Data collected from the flight logs were manually aligned, using the start of the Lissajous trajectory as the reference point. Furthermore, the take-off and landing phases were trimmed off.

Fig. 11 shows the trajectories of all nine experiments for platform A. Overall, the drone can track the reference trajectory with very little overshoot in the Y-axis (± 8 cm). The Z trajectory shows a deviation of 20 cm from the nominal value at the beginning of the flight, which is compensated after the first 5 seconds (time period from -5 seconds to 0 seconds). Then, at time $t = -5$ seconds we switch to "offboard mode," which instead involves using the motion capture system.

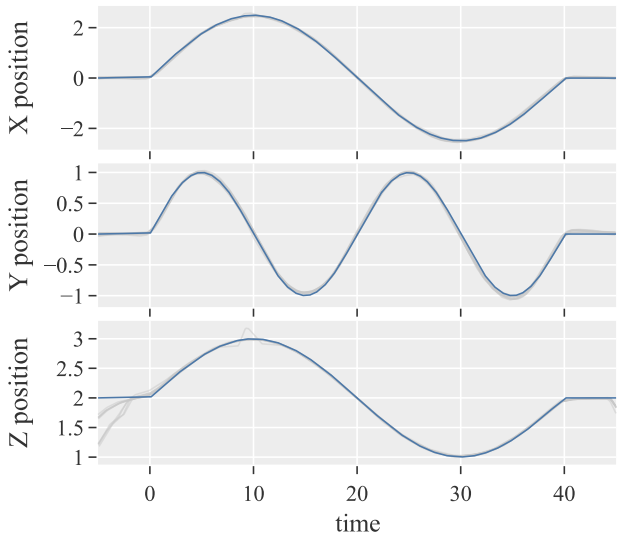

Fig. 11. Trajectories of the positional components (X, Y, Z) vs. time for platform A during nine experiments flight (grey lines) with respect to the setpoint trajectory (blue line). The landing and take-off phases have been trimmed off.

The average RMSE (Fig. 12) provides a more quantitative measure of the tracking performance: the average position error is comparable across the four morphologies, with an average RMSE of 4.3 cm, 4.5 cm, and 1.6 cm for the X, Y, and Z components respectively.

Across the four platforms, C has the worst performance both in terms of mean error and standard deviation. We believe that, since there are only six propulsion modules, this platform might not have enough additional thrust to track at the same time both position and yaw setpoints precisely. This problem and possible mitigations have already been mentioned in Section IV.A and they highlight the need for further development. The solution to this problem would also allow a more robust reconfigurable drone solution in the case of parcels with highly different dimensions in width and length (e.g., transporting, skies, ski poles, pipes).

### B. Outdoor experiments

The two most diverse platforms (A and D) were tested outdoors to validate the drone's performance in a real-world scenario. For simplicity, the drones were programmed with a simple point-to-point navigation mission, using the QGroundControl mission planner, with autonomous take-off and landing at the same location. The flights were performed on agricultural land on a sunny day with a calm wind (gusts up to 9 km/h from SW).

We performed nine outdoor flights for each platform, flying 50 m with a bearing of 16 degrees (from N) at an altitude of 5 m before landing. Batteries were regularly changed to maximize comparability of all flights.

We relied on the onboard RTK GNSS receiver for precise positioning in an outdoor environment. The correction was computed by the ground station computer and sent through the RF telemetry interface. The testing ground was carefully chosen to ensure a line of sight with the drone.

Due to the nature of the outdoor flights (autonomous navigation between waypoints), it is impractical to compute a quantitative measure of the performances of each platform in terms of position error. As such, we only provide the 3D representation of the flight trajectories for both platforms across the nine experiments (Fig. 14).

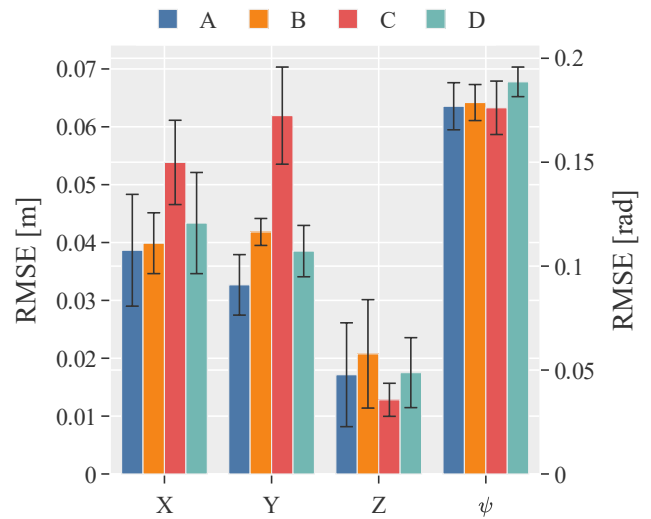

Fig. 12. Average RMSE of the nine experiments: the black whiskers represent the standard deviation. Positional components (X, Y, Z) refer to the leftmost axis, while the yaw refers to the right axis.

While platform A shows very repeatable cruising for all nine experiments, platform D suffered from significant wind disturbance in two flights and generally performed worse. We believe this discrepancy is to blame for the difference in surface area of the platform body, which makes platform D more susceptible to wind. Overall, the performances can be deemed to be appropriate for the targeted application.

## V. CONCLUSIONS AND FUTURE WORK

This work presents a novel approach to transport parcels of different sizes and mass by using a reconfigurable multicopter. The drone is based on a modular design composed of two types of modules: the central computation module and the propulsion module. The propulsion module can be attached to the edges of the parcel through belts. Thus, the parcel becomes the main body of the multicopter. Moreover, we developed a software capable of autonomously computing the modules' positioning and the necessary parameters for the flight controller. A user must only provide the dimensions and mass of the parcel and follow the software's instructions to place the modules around the parcel. The performance of both software and hardware was experimentally tested with two different parcel sizes and two different masses. We performed nine successful indoor flight tests for four drone morphologies using a motion capture system. The tests revealed excellent trajectory tracking, with a very low average position error comparable across the four drone configurations. Additionally, we tested the two most diverse drone configurations available and performed complete autonomous missions from take-off to landing using RTK GPS. These tests confirmed that our system could successfully be used in a real-world scenario.

In the future, we plan to investigate the following aspects to improve the drone performance and reduce the time to prepare the drone to fly: (1) Test the autotuning method proposed by PX4 for fine-tuning the PID gains (http://docs.px4.io/master/en/config/autotune.html) after they are selected from our database; (2) Extend our database by simulating PID gains for the yaw; (3) In the current approach, propulsion modules are symmetrically located on a parcel, considering that the parcel's load distribution is homogenous. We plan to integrate optimization algorithms to find the correct modules configuration when the COG is not in the geometrical center of the parcel; (4) Study the influence of a more aerodynamic shape of parcels on the drone performance; (5) Introduce photo and video tutorials in the software to inform the user where the modules should be placed; (6) Integrate in our software a vision-based algorithm verifying the correct attachment of the modules. Thus, human errors could be minimized.

ACKNOWLEDGMENTS

The authors would like to thank Arthur Gassner, Takuya Kishimoto, Cyrill Lippuner, Noppawit Lertutsahakul, Olexandr Gudozhnik, and Anand Bhaskaran for their work and helpful discussion related to this project.

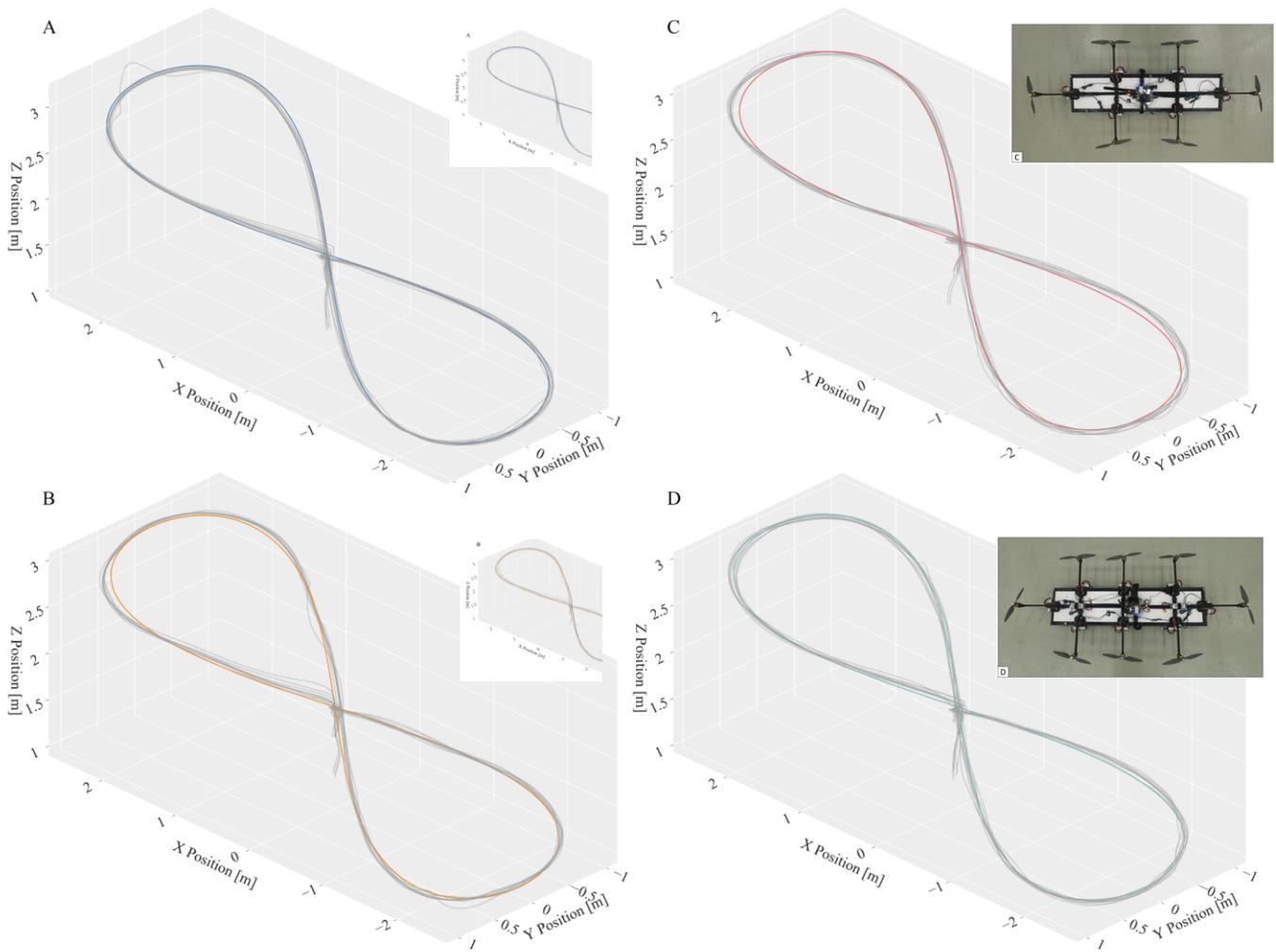

Fig. 13. 3D flight trajectories of platforms A, B, C, and D for the nine INDOOR experiments. The reference trajectory is represented as a colored trace, while each experiment is represented by a gray trace.

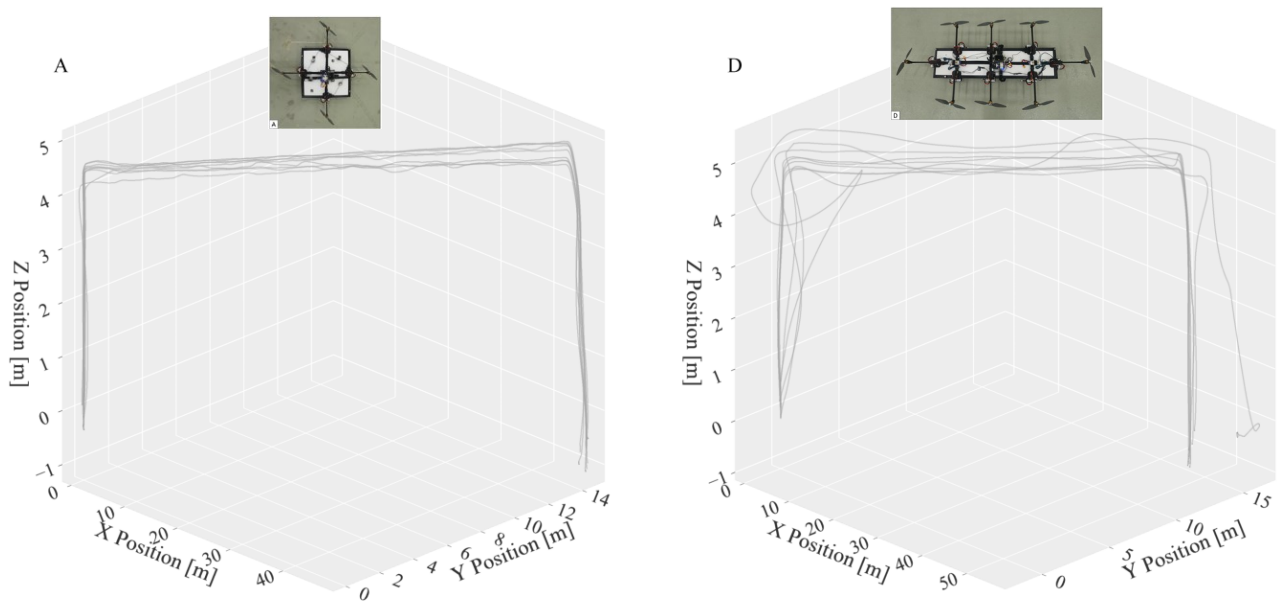

Fig. 14. 3D flight trajectories of platforms A (left) and D (right) for the nine OUTDOOR experiments, each represented by a gray trace. While platform A shows very repeatable cruising for all 9 experiments, platform D did suffer from wind interference in two of the flights.